\documentclass{article}
\usepackage{amsmath,amsfonts,bm}

\def\eqref#1{equation~\ref{#1}}
\def\1{\bm{1}}

\def\rvh{{\mathbf{h}}}

\def\rvs{{\mathbf{s}}}

\DeclareMathAlphabet{\mathsfit}{\encodingdefault}{\sfdefault}{m}{sl}
\SetMathAlphabet{\mathsfit}{bold}{\encodingdefault}{\sfdefault}{bx}{n}
\newcommand{\tens}[1]{\bm{\mathsfit{#1}}}
\def\tA{{\tens{A}}}

\def\gL{{\mathcal{L}}}

\newcommand{\eg}{\textit{e.g.}}

\newcommand{\etens}[1]{\mathsfit{#1}}

\def\etA{{\etens{A}}}

\PassOptionsToPackage{numbers, compress}{natbib}
\usepackage{natbib}

\usepackage[preprint]{neurips_2022}

\usepackage[utf8]{inputenc} % allow utf-8 input
\usepackage[T1]{fontenc}    % use 8-bit T1 fonts
\usepackage{url}            % simple URL typesetting
\usepackage{booktabs}       % professional-quality tables
\usepackage{amsfonts}       % blackboard math symbols
\usepackage{nicefrac}       % compact symbols for 1/2, etc.
\usepackage{microtype}      % microtypography
\usepackage{xcolor}         % colors
\usepackage{enumitem}
\usepackage{multirow}
\usepackage{tabularx}
\usepackage{amsmath}
\usepackage{nicefrac}
\usepackage{adjustbox}
\usepackage{caption}
\usepackage{subcaption}
\usepackage{soul}
\usepackage{wrapfig}

\definecolor{applegreen}{rgb}{0.55, 0.71, 0.0}
\definecolor{aogreen}{rgb}{0.0, 0.5, 0.0}

\definecolor{darker}{rgb}{0,0.15,0.7}
\usepackage[colorlinks, urlcolor=darker, citecolor=darker, linkcolor=darker]{hyperref}

\title{Optimizing Factual Accuracy in Text Generation through Dynamic Knowledge Selection}
\author{Hongjin Qian$^1$,  Zhicheng Dou$^1$, Jiejun Tan$^1$, Haonan Chen$^1$, Haoqi Gu$^2$ \\  \textbf{Xinyu Zhang$^2$, Ruofei Lai$^2$, Zhao Cao$^2$, } and \textbf{Ji-Rong Wen$^1$} \\
$^1$ Gaoling School of Artificial Intelligence, Renmin University of China, Beijing, China \\
$^2$ Huawei Poisson Lab, China \\
\texttt{\{ian,dou\}@ruc.edu.cn}}
\begin{document}

\maketitle

\begin{abstract}
Language models (LMs) have revolutionized the way we interact with information, but they often generate nonfactual text, raising concerns about their reliability. Previous methods use external knowledge as references for text generation to enhance factuality but often struggle with the knowledge mix-up~(\eg, entity mismatch) of irrelevant references. Besides, as the length of the output text grows, the randomness of sampling can escalate, detrimentally impacting the factual accuracy of the generated text. In this paper, we present DKGen, which divides the text generation process into an iterative process. In each iteration, DKGen takes the input query, the previously generated text and a subset of the reference passages as input to generate short text. During the process, the subset is dynamically selected from the full passage set based on their relevance to the previously generated text and the query, largely eliminating the irrelevant references from input. To further enhance DKGen's ability to correctly use these external knowledge, DKGen distills the relevance order of reference passages to the cross-attention distribution of decoder. We train and evaluate DKGen on a large-scale benchmark dataset. Experiment results show that DKGen outperforms all baseline models.

\end{abstract}
 
\section{Introduction}
\label{sec:introduction}
%% Problem

Large language models (LLMs), such as ChatGPT, have shown remarkable capabilities in natural language generation tasks~\citep{radford2019language, t5, brown2020language, smith2022using, ouyang2022training}. Despite this, these generative LMs primarily model the statistical relationships between subword tokens~\citep{sennrich2015neural} and exhibit limited ability in generating factually correct text. Consequently, there is a growing concern regarding the production of nonfactual content~(also called hallucination) by these LLMs~\citep{rae2021scaling, nakano2021webgpt,zhang2022opt, lee2023factuality}. Addressing this issue is crucial for the safe use of such models into real-world applications. 

Many prior studies have aimed to improve the factuality of text generation~\citep{ji2022survey}, and a promising approach among these involves incorporating external knowledge into the text generation process~\citep{lewis2021retrievalaugmented,kgfid, yu2020survey,piktus2021web,west2022probing, shuster2021retrieval}. These techniques generally employ either prepared external knowledge or retrieve knowledge via an information retrieval (IR) system. And the LMs are trained to select and incorporate relevant knowledge from these sources to generate text~~\citep{thoppilan2022lamda, borgeaud2021improving, piktus2021web, petroni2020kilt, guu2020realm}. In industrial settings, the integration of external knowledge to enhance the factual accuracy of text generation has become a popular strategy, as seen in products such as New Bing\footnote{https://news.microsoft.com/the-new-Bing/} and ChatGPT plugins\footnote{https://openai.com/blog/chatgpt-plugins}.

Taking New Bing as an example, it represents a new paradigm in search engines. Here, New Bing takes a user query, extracts reference passages from the top search results, and generates a text as the search answer. To explicitly display the source of the generated answer text, New Bing appends reference marks at the end of the text. These reference marks not only enhance the credibility of the generated answer text but also allow the user to trace back to the source webpage. 

\begin{figure}
    \centering
    \includegraphics[width=\linewidth]{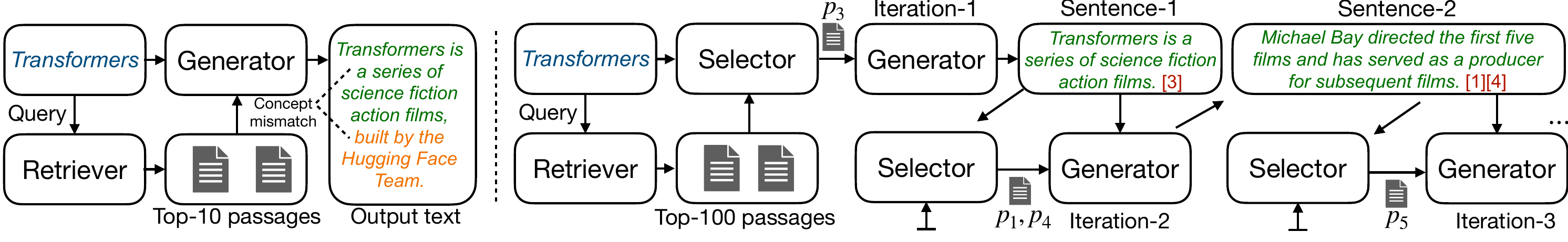}
    \caption{Case comparison of text generation. The left side shows the classic pipeline of retrieval-augmented text generation. The right side shows the iterative text generation proposed in this paper. }
    \label{fig:case}
    \vspace{-10pt}
\end{figure}

While the specific implementation details of New Bing remain undisclosed, such a paradigm is undeniably worth exploring in the research area. However, previous approaches present three primary concerns that may adversely affect the factual accuracy and quality of this New Bing paradigm: \textbf{First}, most of these models simultaneously attend to all reference passages during the generation of each token~\citep{fid20,lakhotia2020fidex,hofstätter2022fidlight}. This design can make the generation model vulnerable to irrelevant or noisy reference passages, which might cause it to blend knowledge from these inappropriate sources~\citep{kgfid}, such as the concept mismatch illustrated in the right side of Figure~\ref{fig:case}. \textbf{Secondly}, these methods are inclined to decode the entire output text in a single pass. As the length of this output text extends, the randomness of the sampling may increase, negatively affecting the factual accuracy of the generated text~\citep{lee2023factuality}. \textbf{Lastly}, most of these methods lack the capacity to generate reference marks as shown in New Bing~\citep{qian2023webbrain}. This deficiency makes it difficult for the end user to accurately locate the source reference passage, thereby undermining reliability.

\textbf{In this paper, we present a model, DKGen, that breaks down the conventional single-pass knowledge-enhanced text generation process into an iterative one.} In each iteration, we dynamically select a subset of knowledge that most accurately supports the generation of the current sentence, given the previously generated sentences. After completing all iterations, we concatenate the sentences generated in each iteration to form the final output text. Importantly, our method enables us to accurately append reference marks at the end of each sentence, enhancing the transparency and reliability of the generated text. Figure~\ref{fig:case}'s right side shows DKGen's iterative generation process. 
%In each iteration, we select a few reference passages from the retrieved results to support text generation.

Specifically, DKGen initiates the text generation process by retrieving a large set of reference passages (\eg, top-100) via an Information Retrieval (IR) system using the input query. In each iterative step of generation, DKGen re-ranks these reference passages based on their relevance to both the previously generated text and the input query (detailed in Section~\ref{sec:dk}). Next, DKGen uses the previously generated text, the selected relevant passages, and the query as inputs to generate a new sentence (discussed in Section~\ref{sec:generation}). In each iteration, DKGen only generates a short text (\eg, a sentence) using carefully selected reference passages. This approach significantly mitigates issues such as knowledge mix-up from irrelevant reference passages and growing sampling randomness, thereby enhancing the factuality of the generated text. Moreover, DKGen incorporates the relevance scores as an order of importance to further augment its capacity to correctly utilize knowledge from the reference passages. This is achieved by distilling the relevance scores into the cross-attention distribution of DKGen's decoder (elaborated in Section~\ref{sec:distill}). Importantly, DKGen's iterative generation does not reduce decoding efficiency, as its decoder only needs to attend to a limited number of reference passages and decode a short text in each iteration. Furthermore, we encode the reference passages only once to avoid redundant computations (analyzed in Section~\ref{sec:infer}).

In this paper, we train and evaluate DKGen on a public benchmark dataset known as WebBrain~\citep{qian2023webbrain}, which was constructed by crawling the English Wikipedia and the corresponding reference articles. To facilitate DKGen's iterative sentence generation training, we partition the WebBrain training dataset into sentence-level granularity (details provided in Section~\ref{sec:dataset}). In our experiments, we compare DKGen against competitive models, including ChatGPT, and find that DKGen outperforms all baseline models.
The contributions of this study are threefold:
(1)~We introduce an iterative text generation method, which allows more flexibility of incorporating external knowledge during the generation process. This method can be applied to a wide range of existing text generation models.
(2)~We adapt a dataset from a public benchmark dataset to facilitate the training of iterative text generation.
(3)~We introduce DKGen, a model implementing the iterative text generation method. By dynamically selecting knowledge for each sentence generation, DKGen enhances the factuality of text generation, facilitating the precise assignment of reference marks to individual sentences.

\section{Related Work}
\label{sec:related_work}

\paragraph{Factuality in Text Generation}
Pre-trained language models often generate hallucinations, or factually incorrect statements~\citep{krathwohl2002revision, petroni2019language, aspillaga2021inspecting, zhou2020evaluating}. Various methods have been proposed to improve the factuality of text generation: (1)~studies such as \citep{lee2023factuality, petroni2019language, aspillaga2021inspecting, zhou2020evaluating} suggest that larger language models are better equipped to recall knowledge from an extensive training corpus, improving their factual accuracy. (2) some researchers seek to rectify factual errors in the post-processing stage~\citep{de2021editing,jang2021towards,meng2022locating}.
(3)~many downstream tasks utilize task-specific language models that have been fine-tuned for a range of text generation tasks, such as summarization, machine translation, and dialogue systems~\citep{cao2018faithful,dong2020multi,huang2020knowledge,huang2021factual,cao2021cliff,zhu2021enhancing,chen2021improving,wiseman2017challenges,nie2019simple,liu2021towards,su2021plan,wang2021sketch,rebuffel2022controlling,shen2021identifying,shuster2021retrieval,rashkin2021increasing,wu2021controllable,dziri2021neural}; (4) a subset of studies focus on addressing factual errors in the parametric knowledge of language models, which is derived from the training corpus~\citep{jiang2020can, zhong2021factual,elazar2021measuring}.(5) human feedback or demonstrations have been found to be invaluable in improving the factual accuracy of language models. For instance, InstructGPT was fine-tuned using collected human feedback to enhance truthful generation~\citep{ouyang2022training}. WebGPT was trained to cite its sources when generating output, enabling humans to verify factual accuracy by checking whether a claim is backed by a reliable source~\citep{nakano2021webgpt}. Many of these methods utilize external knowledge to enhance the factuality of text generation, which is considered as a promising and effective way to reduce hallucinations.

\paragraph{Knowledge-Augmented Text Generation}
Retrieval-augmented text generation is being viewed as a promising solution to counteract hallucinations~\cite{li2023apibank, lewis2021retrievalaugmented, qian2023webbrain,fid20}. This technique has been widely implemented in various NLP tasks such as machine translation~\citep{DBLP:conf/acl/00020LLL20}, open-domain question answering~\citep{lewis2021retrievalaugmented,guu2020realm}, dialogue generation~\citep{DBLP:journals/ir/ZhuDNW20, glaese2022improving} and open-ended text generation~\citep{lee2023factuality,qian2023webbrain}. Most of these approaches produce answers by retrieving passages as reference from either a retriever or search engine, citing the reference passages to support text generation.
For instance, \citet{menick2022teaching} employ reinforcement learning via human preferences to train LLMs that can generate answers and cite relevant evidence to support their claims. In contrast, WebGPT~\citep{nakano2021webgpt} finetunes GPT-3~\citep{brown2020language} to search and browse the internet for pertinent information. LaMDA~\citep{thoppilan2022lamda} enhances factual grounding and safety by utilizing annotated data for training and empowering the model to consult external knowledge sources. \citet{glaese2022improving} employ various sources of evidence and guidelines to create dialogue systems. However, most of these methods remain susceptible to irrelevant knowledge from reference passages, which may cause them to blend information from different sources. In this paper, we tackle this issue by carefully selecting reference passages while generating each sentence, thereby increasing the credibility of the knowledge source.

\section{Methodology}

% In this section, we first formalize the problem of generating factual article with reference mark in section \ref{sec:formal}. Then, we givr a short recap of the Fusion-in-Decoder model and their variants, which are widely used in the domain of knowledge-enhanced text generation (in \ref{sec:fid}). 

% In section \ref{sec:overview}, we give an overview of the proposed model and in the following section, we introduce how we select useful reference considering its relevance to both context and the main topic.
\subsection{Preliminary}
The problem of generating text using external knowledge can be defined as finding the text $T^*$ that maximizes the conditional probability given an input query $q$ and a knowledge prior $\Theta$. Formally, we can express this as:
\begin{align}
\label{eq:overall_def}
T^* = \arg\max_{T} \texttt{Prob}(T | q, \Theta),
\end{align}
where $\texttt{Prob}(T | q, \Theta)$ represents the probability of generating text $T$ given the input query $q$ and the knowledge prior $\Theta$. The knowledge prior $\Theta$ is a combination of parametric knowledge stored inside the language model and external knowledge obtained from other sources such as search engines or knowledge corpora. Here, we use a reference passage set $P$ as the external knowledge, which comprises of a set of passages $(p_1, \ldots, p_k)$ that can be used to support text generation.

There are two types of method for incorporating $P$ into the text generation process: (1) for decoder-only models, the reference passages can be considered an essential component of the input sequence, along with the input query (e.g., the prompt)~\cite{brown2020language}; (2) for encoder-decoder models, the reference passages can be processed either jointly or separately within the encoder, allowing the decoder to utilize the reference passages by attending to the encoder's hidden states~\cite{fid20,lewis2019bart}. Both types of models generate the output text $T = (t_1, \ldots, t_n)$ by accounting for the conditional probability $p(t_n|P, t_{1:n-1})$, where $t$ represents a token. 

Many of these methods present two primary concerns that may adversely impact factuality of the generated text: (1)~the decoder jointly attends to all reference passages while generating each token $t$. This renders the decoder susceptible to irrelevant or noisy reference passages, leading it to potentially blend knowledge from such irrelevant sources, such as entities~\cite{kgfid}; (2)~these method tend to decode the full output text in a single pass. As the length of the output text increases, the randomness of sampling can escalate, detrimentally impacting the factual accuracy of the generated text~\cite{lee2023factuality}.

\subsection{Problem Formalization} \label{sec:formal}

\textbf{In this paper, we propose dividing the process of generating the entire text in one pass into an iterative process to solve the above mentioned concerns.} We initialize the process by retrieving a set of reference passage $P=(p_1,\cdots\,p_k)$ using the input query $q$. In each generation iteration, we dynamically select a subset of reference passage $P_i\in P$ to support partial text generation (e.g., a sentence $s_i$). Formally, we define the iterative text generation process as:
\begin{align}
\label{eq:def}
s_i^* = \arg\max_{s_i} p(s_i | q, P_i, T_{1:i-1}), \quad T = (s_1^{\gets{P_1}}, \cdots, s_m^{\gets{P_m}}),
\end{align}
where the generated text $T$ consists of $m$ sentences $s_i, i\in[1,m]$. The term $s_i^{\gets{P_i}}$ indicates that the sentence $s_i$ is supported by a subset of reference passage $P_i$ which is dynamically selected
%The main feature of the subset prior knowledge $\theta_i$ is that it selects only a small number of reference passages $P_i\in P$ to support text generation. 
%The selected reference passages $P_i$ change 
when generating different sentences $s_i$ by considering the previously generated text $T_{1:i-1}$ and the given query $q$. Formally, we have:
\begin{align}
\label{eq:prob}
    P^*_i=\arg\max_{P_i\in P} \texttt{Prob}(P_i|T_{1:i-1},q),
\end{align}
where $P^*_i$ refers to the most proper reference passage set that can support generating the sentence $s_i$.

By the definitions provided in Eq. (\ref{eq:def}) and Eq. (\ref{eq:prob}), the factual accuracy of the generated text can be improved by: (1) shortening the target text during generation~(\eg, sentence-level) to increase the determinacy of the sampling process; (2) supplying more precise reference passages for each sentence generation to alleviate the knowledge mix-up issue; and (3) explicitly knowing the cited reference passages for each generated sentence.

% Consequently, the main problems to address in this paper are: (1) the design of an effective method for generating the text $T$ sentence by sentence while maintaining coherence, and (2) the selection of an appropriate reference passage set $P_i$ that can better support the generation of the sentence $s_i$.

\subsection{The proposed model: DKGen} \label{sec:overview}

\begin{figure}
    \centering
    \includegraphics[width=0.8\linewidth]{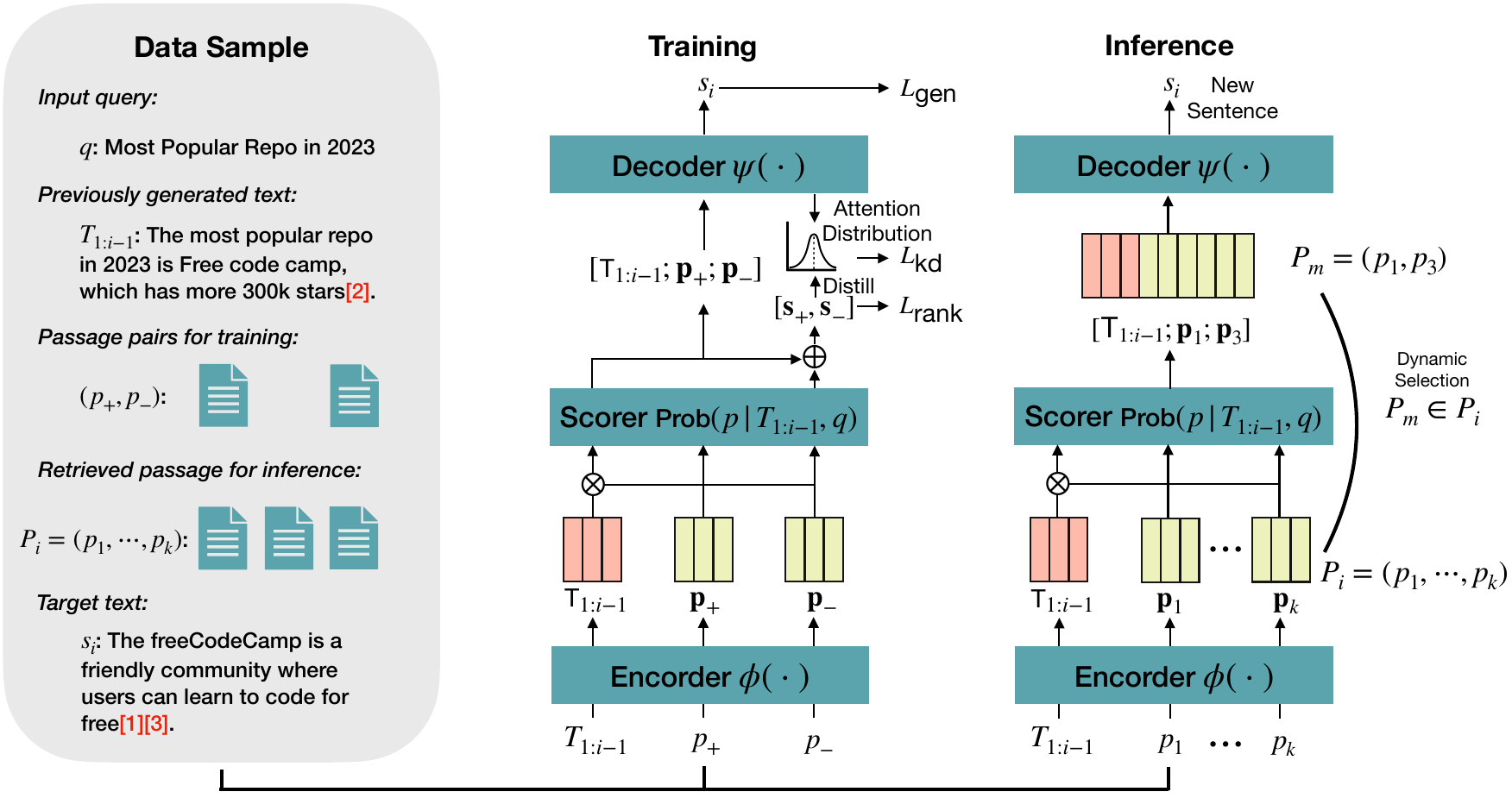}
    \caption{Overview of the Proposed Model: DKGen}
    \label{fig:overview}
    \vspace{-10pt}
\end{figure}

To iteratively generate the text $T=(s_1,\cdots,s_m)$ sentence by sentence, intuitively, we can halt the decoding process upon generating a complete sentence (e.g., $s_{i-1}$). Subsequently, the previously generated text $T_{1:i-1}=(s_1,\cdots,s_{i-1})$, the newly selected reference passages $P_i$, and the input query $q$ can be incorporated to produce a new sentence $s_i$. This paper presents the DKGen model, which adheres to this intuition while enhancing factual accuracy and text fluency through specialized model designs. Figure \ref{fig:overview} provides an overview of DKGen.

The training procedure of DKGen (illustrated on the left side of Figure \ref{fig:overview}) optimizes it by minimizing the discrepancy between a hypothesis sentence $s_i\sim \psi(T_{1:i-1}, P_i, q)$ and a ground-truth sentence $s^*_i$, where $\psi(\cdot)$ denotes a decoder. In the process, three training losses are optimized: (1) the text generation loss $\mathcal{L}_{\texttt{gen}}$, wherein the ground-truth sentence $s^*_i$ serves as the target text label~(in section~\ref{sec:generation}); (2) a ranking loss $\mathcal{L}_{\texttt{rank}}$ that assists the model in identifying appropriate reference passages to better support text generation~(in section~\ref{sec:dk}); and (3) a knowledge distillation loss $\mathcal{L}_{\texttt{kd}}$, which employs ranking scores as soft labels to supervise the decoder's attention distributions across different input passages~(in section~\ref{sec:distill}).

During the inference process (depicted on the right side of Figure \ref{fig:overview}), DKGen generates a new sentence $s_i$ by first employing a scoring model, optimized by $\mathcal{L}_\texttt{rank}$, to evaluate all reference passages and retain a subset $P_i\in P$ as the supporting passage for the current generation iteration. Then, DKGen incorporates the previously generated text $T_{1:i-1}$, the selected reference passage $P_i$, and the input query $q$ to generate the new sentence $s_i$. After generating all sentences $T=(s_1, \cdots, s_m)$, we can get the final output text by combine the $m$ sentences, and attach the cited reference marks at the end of each sentence~(in section~\ref{sec:infer}). The following sections will discuss the details of DKGen.

\subsection{Iterative Sentence Generation} \label{sec:generation}
DKGen employs an encoder-decoder model~(\eg, BART or T5)~\cite{lewis2019bart,t5} as the underlying language model to perform text generation. By the Eq. (\ref{eq:def}), we take the query $q$, the previously generated text $T_{1:i-1}$, and the selected reference passage set $P_i$~(we will discuss how to select $P_i$ in section \ref{sec:dk}) as input to generate $s_i$. Supposing $P_i$ has $j$ reference passages, we separately process $T_{1:i-1}$ and the $j$ reference passages into text sequences. Taking the $j$-th reference passage $p_j$ as an example:
\begin{align}
    \label{eq:passage}
    p'_j=\texttt{[query]} q \texttt{[ref]} p_j\texttt{[EOS]}, \quad T'_{1:i-1}=\texttt{[context]} T_{1:i-1}\texttt{[EOS]},
\end{align}
where $\texttt{[query]}$, $\texttt{[ref]}$, $\texttt{[context]}$ and $\texttt{[EOS]}$ are special tokens, which are used to help the model to distinguish different parts of the input text sequences.

Afterwards, we encode the $j$ sequences $(p'_1, \cdots, p'_j)$ and the sequence $T'_{1:i-1}$ separately to obtain their corresponding hidden states $(\rvh_1,\cdots,\rvh_j)$ and $\mathbf{T}_{1:i-1}$ using an encoder $\phi(\cdot)$:
\begin{align}
\label{eq:encoder}
    \rvh_j = \phi(p'_j), \quad \mathbf{T}_{1:i-1}=\phi(T'_{1:i-1}).
\end{align}
%% TODO
And then, the decoder $\psi(\cdot)$ attends over the concatenation of the hidden states of all the sequences, and outputs a text sequence $s_i$ with auto-regressive mechanism:
\begin{align}
    s^* \sim \psi(s^*,\rvh) = \prod_{n=1}^{|s^*|}p(s^*|s^*_{<n},\rvh), \quad \rvh = [\mathbf{T}_{1:i-1};\rvh_1;\cdots;\rvh_j].
\end{align}
After obtaining sentence $s_i$, we append $s_i$ into $T_{1:i-1}$ to get $T_{1:i}$. Incorporating with query $q$ and a new reference passage set $P_{i+1}$, we can generate the next sentence $s_{i+1}$ with the same process above. 
We optimize the sentence generation by minimizing the negative log-likelihood function of the ground-truth text:
\begin{align}
    \gL_{\texttt{gen}}=\gL_{\text{NLL}}(x,s^*) = -\sum_{n=1}^{|s^*|}\log p(s^*_n|s^*_{<n},x).
\end{align} 

\subsection{Dynamic Knowledge Selection for Sentence Generation} \label{sec:dk}
In the process of iterative sentence generation, it is crucial to select the proper reference passage subset $P_i\in P$ to support generating sentence $s_i$. Following the Eq. (\ref{eq:prob}), DKGen selects the subset $P_i$ from the full set $P$ by considering the the relevance $\texttt{Prob}(p|T_{1:i-1},q)$ of each passage $p\in P$. To do this, we need to compute the relevance for each passage $p\in P$. To be concrete, DKGen separately measures the matching degree of $(p,q)$ and $(p,T_{1:i-1})$. For each $p\in P$ we have:
\begin{align}
\label{eq:two_part}
     \texttt{Prob}(p|T_{1:i-1},q) \sim \texttt{Prob}(p|q)+ \texttt{Prob}(p|T_{1:i-1}).
\end{align}
\textbf{First}, we compute $\texttt{Prob}(p|q)$ for each reference passage $p\in P=(p_1,\cdots,p_k)$.
%, we separately measure the matching degree $\texttt{Prob}(p|q)$ of each reference passage. 
Recall that in Eq.~(\ref{eq:passage}) and Eq.~(\ref{eq:encoder}), we concatenate the query $q$ with a reference passage $p$ into a text sequence $p'$, which is then fed into the encoder $\phi(\cdot)$ to obtain its hidden states. This process mirrors the ranking task using a cross-encoder~(note that the encoder of BART or T5 is bi-directional)~\cite{humeau2020polyencoders}, which also aims to compute the relevance of a query-passage pair $(q,p)$. Therefore, DKGen directly uses the hidden states output by Eq.~(\ref{eq:encoder}) to compute $\texttt{Prob}(p|q)$.
%Taking the $k$-th reference passage $p_k\in P$ as an example,
%we can separately measure the matching degree $\texttt{Prob}(p_k|q)$ of each reference passage $p_k\in P$ with query $q$. 

Specifically, taking the $k$-th reference passage $p_k$ as an example, we pool its sequence hidden states $\rvh_k$ by extracting the hiden states of the $\texttt{[EOS]}$ token~\footnote{Using $\texttt{[EOS]}$ to represent the sequence  follows the original BART paper~\cite{lewis2019bart}.} of the sequence $p'_k$ by: $\hat{\rvh}_k = \texttt{Pool}_{\texttt{[EOS]}}(\rvh_k)$. For all $k$ reference passage, we have $\hat{\rvh} = (\hat{\rvh}_1,\cdots,\hat{\rvh}_k)$.
We then utilize a multi-layer perceptron~(MLP) to map $\hat{\rvh}$ into matching scores, and compute the $\texttt{Prob}(P|q)$ using a $\texttt{Softmax}$ function:
\begin{align}\label{eq:score1}
    \texttt{Prob}(P|q) = \texttt{Softmax}(\texttt{MLP}(\hat{\rvh})).
\end{align}
\textbf{Second}, we compute $\texttt{Prob}(p|T_{1:i-1})$. Given $P=(p_1,\cdots,p_k)$ and the previously generated text $T_{1:i-1}$, we have their pooled sequence representation $\hat{\rvh} = (\hat{\rvh}_1,\cdots,\hat{\rvh}_k)$ and $\hat{\mathbf{T}}_{1:i-1}$. Then, we compute $\texttt{Prob}(P|T_{1:i-1})$ by:
\begin{align}
\label{eq:context}
    \texttt{Prob}(P|T_{1:i-1}) = \texttt{Softmax}(\hat{\rvh} \cdot \hat{\mathbf{T}}_{1:i-1} / \sqrt{d}),
\end{align}
where $d$ is the dimension size. Eq.~(\ref{eq:context}) presents a simple yet effective measurement for the $T_{1:i-1}$-aware reference ranking. Using it we can filter the references that are not topically consistent with the previously generated text $T_{1:i-1}$, and therefore guarantee the topical coherence of sentences in the output text.
Afterwards, we can get the reference passage subset $P_i$ by selecting the reference passages that have the $j$ biggest relevance score $\rvs = \texttt{Prob}(p|q)+\texttt{Prob}(p|T_{1:i-1})$. 

During the training phase, we optimize the $\texttt{Prob}(p|T_{1:i-1},q)$ using a Pairwise Ranking Loss~$\gL_{\texttt{rank}}$, in which we use the supporting reference passage $p_+$ as positive passage and sample a negative passage $p_-$ from the corpora~(see Section~\ref{sec:dataset}).

\subsection{Attend to the Useful Reference through Relevance Distillation}
\label{sec:distill}
In Section~\ref{sec:dk}, we discuss how to select a proper reference subset $P_i\in P$ to
support generating a sentence $s_i$. The selection process mainly consider the relevance $\texttt{Prob}(p|T_{1:i-1},q)$, with which we can get $P_i$ by selecting reference passages with high relevance scores. This indicates that reference passages in $P_i$ have an importance order. Reference passage with a high score is more likely to provide useful knowledge for text generation, which implies that when decoding the output text, the decoder should pay more attention to the reference passages with higher scores. Inspired by this intuition, we design a relevance distillation mechanism to supervise the decoder's attention distributions with the relevance scores of input passages.

Taking the $j$-th reference passage $p_j\in P_i$ as an example, we can extract $p_j$'s cross-attention matrix $\tA_j\in \mathbb{R}^{n_1 \times n_2 \times l_1 \times l_2}$ from the decoder $\psi(\cdot)$, where $n_1$ and $n_2$ refer to the number of transformer layers and the number of attention heads stacked in the decoder. $l_1$ and $l_2$ are the token length of reference passage $p_j$ and the output text, respectively. The details for computing $\tA_j$ can refer to~\citet{vaswani2017attention}.
We then average $\tA_j$ over all dimensions and use a $\texttt{Softmax}$ function to get the attention score distribution $\rvs_{\texttt{att}}$ of  all of the $j$ reference passages $P_i=[p_1,\cdots,p_j]$:
\begin{align}
    \rvs_{\texttt{att}} = \texttt{Softmax}([\etA_1, \cdots, \etA_j]), \quad \etA_j = \mathop{{\texttt{Pool}_\texttt{ave}}}\limits_{n1,n2,l1,l2}(\tA_j).
\end{align} 
And we use the conditional probability $\texttt{Prob}(p|T_{1:i-1},q)$ computed in Section~\ref{sec:dk} as the reference scores, which we denote as $\rvs_{\texttt{rel}}$, to supervise the attention score distribution. We use a KL divergence loss to optimize the relevance distillation:
\begin{align}
    \gL_{\texttt{kd}} = \rvs_{\texttt{rel}}^\top \cdot (\log \rvs_{\texttt{rel}} - \rvs_{\texttt{att}})
\end{align}
\textbf{Traing Loss}: During the training phase, we optimize the three loss function together by:
\begin{align}
    \gL = \alpha\gL_{\texttt{gen}} + (1-\alpha)(\gL_{\texttt{rank}} + \gL_{\texttt{kd}}),
\end{align}
where $\gL$ is the overall loss function and $\alpha$ is a scalar.

\subsection{Inference: Sentence Decoding with Dynamic Reference Selection}
\label{sec:infer}
DKGen initializes inference by retrieving $k$ reference passages, denoted as $P=[p_1,\cdots,p_k]$, from a large passage corpora based on the input query $q$. In this study, we employ stand-alone retrievers for this purpose, and the influence of different retrievers is discussed in Section~\ref{sec:retriever}. Subsequently, DKGen calculates the query-aware relevance, represented as $\texttt{Prob}(p|q)$, for each passage $p\in P$ using the formulation presented in Eq.~(\ref{eq:score1}). 
In our experiments, instead of actively retrieving reference passages at each iteration, we remain the reference passage set $P$ fixed throughout the generation process, mainly considering inference efficiency.
As the reference passage set is fixed, there is no need to compute $\texttt{Prob}(p|q)$ repeatedly in subsequent generation iterations. 
At each iteration, DKGen additionally computes $\texttt{Prob}(p|T_{1:i-1})$ as per Eq.~(\ref{eq:context}). The reference passages in $P$ are then sorted based on the score $\texttt{Prob}(p|q)+\texttt{Prob}(p|T_{1:i-1})$, and the top-$m$ passages are selected to form a subset $P_m$. This subset is utilized to facilitate the generation of the sentence $s_i$, as described in Section~\ref{sec:generation}.

\textbf{Decoding Efficiency Analysis} One potential question regarding DKGen pertains to the efficiency of iteration generation comparing to decoding all text in a single pass. In practice, the iteration generation would not be slower. The reasons are: Firstly, the time complexity of the transformers decoder is approximately $O(n^2)$, where $n$ represents the length of the input. Instead of inputting all $k$ retrieved passages into the decoder, DKGen selectively chooses $m$ passages as input (e.g., with $k=100$ and $m=2$). This significantly reduces the computational burden associated with the decoder's cross-attention. Secondly, typical text decoding strategies, such as Beam Search, exhibit a time complexity of approximately $O(k^n)$, where $k$ denotes the beam width and $n$ represents the length of the decoded text. DKGen, on the other hand, decodes only a short text at each iteration, resulting in significantly faster processing compared to decoding longer texts. In Table~\ref{tab:overall}, we evaluating the per query latency of all models, verifying that DKGen is relatively more efficient than baselines.

\section{Experiments}
\begin{table*}[t]
    \centering
    \small
    \caption{Performance of all models. The best results are in bold. Latency is measured in (ms/query).}
    % \begin{center}
    \begin{tabular}{lccccccccc}
    \toprule
        Model &  BLEU-1 & BLEU-4  & ROUGE-L & BARTScore & FactScore & TripleScore & Latency   \\
    \midrule
        BART    &   14.52  & 2.33 & 13.57 & -4.855 & 7.01 & 13.02  & 1125\\
        BART-L  &   18.61  & 5.02 & 17.45 & -4.731 & 7.49 & 13.82  & 1788\\
        FiD     &   23.95  & 5.86 & 18.61 & -4.502 & 9.87 & 18.39  & 879\\
        FiD-L   &   25.91  & 6.01 & 20.64 & -4.414 & 10.56 & 19.56 & 1453\\

        ReGen &     24.70  & 7.48 & 21.06 & -4.475 & 12.08 & 20.43  & 1096\\
        ReGen-L&    27.81  & 7.92 & 22.00 & -4.303 & 12.25 & 20.33  & 1565\\
        \midrule
        ChatGPT & 24.22 & 5.44 & 19.56 & \textbf{-4.280} & 10.64 & 19.21 & -\\
        ChatGPT-R & 25.19  & 5.79 & 19.35 & -4.320 & 11.98 & 22.87 & -\\
        \midrule
        DKGen   &   30.77  & 8.95 & 23.29 & -4.473  &  12.84 & 24.76 & \textbf{531} \\
        DKGen-L &   \textbf{31.48}  & \textbf{9.26} & \textbf{23.57} &  -4.452& \textbf{13.01} & \textbf{26.43} & 1310\\
        
    \bottomrule
    \label{tab:overall}
    \end{tabular}
    \vspace{-15pt}
    % \end{center}
\end{table*}
\subsection{Dataset and Evaluation Metrics}\label{sec:dataset}

We conduct experiments on the WebBrain dataset~\citep{qian2023webbrain}, a benchmark dataset specifically designed for generating factually correct text based on large web corpora. We use WebBrain-G to train all baselines, and WebBrain-R to train the retriever model employed in this paper. Each data sample in WebBrain-G consists of a query $q$, a reference passage set $P$, and the target text $T=(s_1, \cdots, s_m)$. To train DKGen, we split the target text into sentences and construct training data samples as $(q, p_+, p_-, T_{1:i-1}, s_i)$, where the positive passage $p_+ \in P$ refers to the reference passage supporting the target sentence $s_i$, and the negative passage $p_- \in P$ is randomly sampled from $P$. The training objective shifts from generating the target text $T$ to a sentence $s_i\in T$. For more details of WebBrain, please refer to \citet{qian2023webbrain}.

\textbf{BLEU}~\citep{bleu}, and \textbf{ROUGE}~\citep{rouge} are metrics employed to measure the $n$-gram overlap between the generated text and the ground-truth text. Higher values of these metrics denote a greater similarity between the generated and ground-truth text. We utilized the \texttt{nlg-eval} package~\citep{nlg-eval} to compute these metrics. Given the recent advancements in model-based metrics capable of assessing semantic-level text similarity, we also utilize \textbf{BARTScore}~\citep{bartscore} as part of our evaluation methodology.\footnote{We use the checkpoint provided in the Github repository \url{https://github.com/neulab/BARTScore}.}

To evaluate the factuality, we use \textbf{TripleScore}~\citep{triple-score} and \textbf{FactScore}~\citep{factscore}. The Triple Score is computed by first extracting lexical components from the source text and the generated text to form semantic triplets using OpenIE~\citep{openie}. The score is then calculated based on the precision of the semantic triplets in the generated text, with higher scores indicating more accurately generated semantic relations. The FactScore is computed by first extracting the entity-level relation of each sentence, and then it is calculated based on the precision of the entity-level relation in the generated text. Higher FactScore indicates better entity-relation factuality. 
Both FactScore and Triple Score were computed using the \texttt{FactSumm} package~\citep{factsumm}. Note that we combine the target text and reference passages as the source text, which implies that, being supported by either the target text or the reference passages, the generated text would be considered as factually correct.

\subsection{Experimental Settings}
DKGen employs BART as its foundational language model. For comparison, we also report the performance of the vanilla BART~\citep{lewis2019bart}, FiD~\citep{fid20}, and ReGen~\citep{qian2023webbrain}. Each of these models is initialized using off-the-shelf checkpoints provided by HuggingFace~\citep{huggingface}. Both base and large models are trained for all baselines, with the large models denoted by the suffix "-L" (e.g., DKGen-L) for clarity. We also evaluate the performance of ChatGPT using OpenAI's APIs. After comparing the response quality, we find the prompt ``\texttt{introduce [X] in Wikipedia-style}'' yields the most satisfactory results. Furthermore, we experiment with providing the retrieved reference passages to ChatGPT, using the prompt ``\texttt{introduce [X] in Wikipedia-style with the references:[references]}''. We denote this variant as ChatGPT-R. We use the SPLADE~\citep{splade} model as the retriever, which is trained by WebBrain-R. We will compare the impact of different retriever model in section~\ref{sec:retriever}.

Training of all baseline models was conducted using 8 Tesla V100 32G GPUs. The batch size per GPU was configured to the maximum number that wouldn't exceed the available GPU memory. For instance, the per GPU batch sizes for DKGen and DKGen-L were set to 56 and 16 respectively. We train all models for 5 epochs. The AdamW optimizer~\citep{adamw} was used for optimization, with a learning rate of 5e-5.
During the inference stage, 20 reference passages were retrieved from the full passage corpus of WebBrain~(204M passages) as the external knowledge. For baseline models that incorporate all reference passages as input, we supplied them with the top-5 reference passages and set the maximum generation length to 256. For DKGen, we selected either 1 or 2 reference passages in each iteration, limiting the maximum sentence generation length to 64 (if the relevance score of the 2nd passage exceeded a threshold $\gamma>0.8$, we retained both reference passages). The iterative process would stop once 5 reference passages had been utilized.

\paragraph{Main Results} \label{sec:result}
The overall results of the experiment are presented in Table \ref{tab:overall}, from which several key findings can be derived: (1) Regarding the $n$-gram overlapping metrics  BLEU and ROUGE, DKGen surpasses all baseline models, suggesting that it can generate text with better lexical quality; (2) Regarding the model-based metric BARTScore, DKGen falls short compared to ChatGPT and ReGen-L, indicating that there is still room for improvement in the semantic quality of DKGen (\eg, fluency and coherence); (3) Regarding factuality metrics, DKGen outperforms all baselines, which confirms the effectiveness of DKGen's model design in enhancing the factuality of the generated text. 
\subsection{Discussions}

\paragraph{Ablation Study}
\begin{wrapfigure}[13]{R}{0.5\textwidth}
\vspace{-10pt}
\centering
\includegraphics[width=\linewidth]{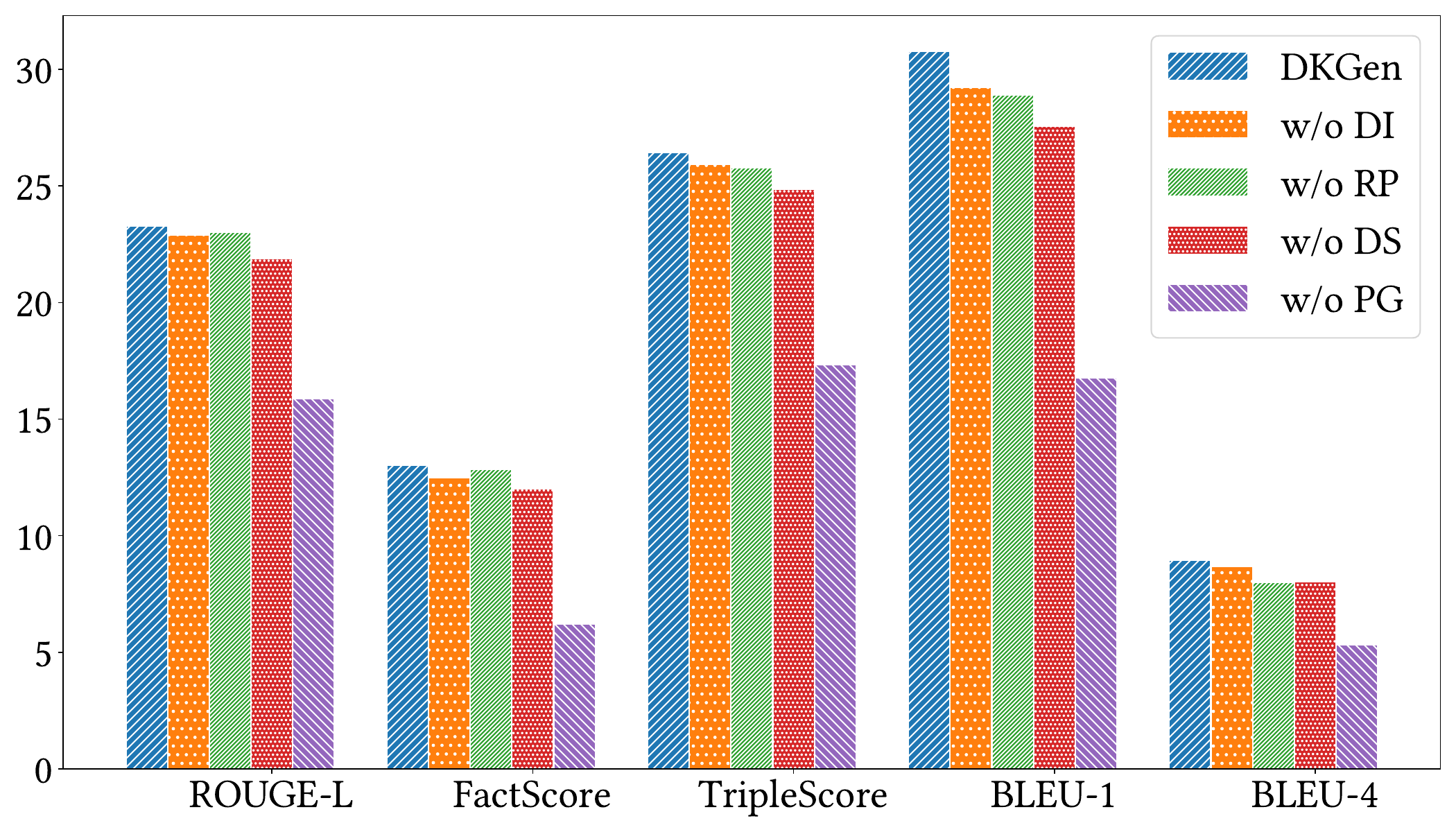}
\caption{Results of Ablation Studies\label{fig:abl}}
\end{wrapfigure}
To evaluate the effectiveness of DKGen's model components, we perform the following ablation studies: (1) w/o DS: integrating all reference passages into DKGen without utilizing \textbf{D}ynamic \textbf{S}election; (2) w/o DI: eliminating the knowledge \textbf{DI}stillation loss, $\gL_{kd}$, from the training process; (3) w/o PG: excluding the \textbf{P}reviously \textbf{G}enerated text, $T_{1,i-1}$, from the model input during inference; (4) w/o RP: discarding the \textbf{R}elevance between the \textbf{P}reviously generated text and the reference passages, $\texttt{Prob}(P|T_{1:i-1})$, during inference. Figure~\ref{fig:abl} presents the results of these ablation studies, from which we can draw the following conclusions: (1) the removal of any DKGen component results lead to a decline in performance, thereby validating the effectiveness of the DKGen's model designs; (2) the largest performance drop is observed when the previously generated text is omitted, emphasizing its significance in generating subsequent sentences; (3) excluding dynamic knowledge selection also leads to a substantial decrement in performance, indicating that knowledge selection is essential for knowledge-enhanced text generation.

\begin{wraptable}{r}{7cm}
\centering
\vspace{-4.5mm}
\small
\caption{Impact of different retrievers.\label{tab:ret_selected}}
\vspace{-5pt}
\begin{tabular}{cccc}\\\toprule  
\textbf{Retrievers} & BM25 & DPR\citep{xiao2022retromae}  & SPLADE\citep{splade}\\ \midrule
ROUGE-L &22.01 & 21.56 & \textbf{23.29}\\  
FactScore & 12.06 & 11.90 & \textbf{12.84}\\
TripleScore & 24.29 & 23.88 & \textbf{24.76} \\
\bottomrule
\end{tabular}
\vspace{-7pt}
\end{wraptable} 
\vspace{-5pt}
\paragraph{Impact of Retriever} \label{sec:retriever}
In this paper, our primary focus is not on developing improved retrievers for knowledge-enhanced text generation. However, the performance of retrievers has a direct impact on the final generation outcome. Thus, we investigate the influence of various retrievers, with the results displayed in Table~\ref{tab:ret_selected}. These findings suggest that the SPLADE retriever is more effective in bolstering the generator's overall performance. Consequently, we conduct our experiments using the retrieval results from SPLADE. Furthermore, we believe that the development of a retriever that takes into account both the query and the previously generated text could enhance DKGen's generation performance. This remains an area for future exploration.
\vspace{-5pt}
\paragraph{Case Study}
\begin{table*}[t]
\small
    \centering
    \caption{Comparisons of text generated by DKGen and ChatGPT. Text in green denotes factual consistency with the target text or reference passages, while text in orange refers to contradictions.}
    \begin{tabular}{p{.98\linewidth}}
    \toprule
     \textbf{Query}: Durg district \\
     \textbf{Ground-truth target}:  Durg district is a district situated in Chhattisgarh, India. The district headquarters is Durg. The district covers an area of 2,238 km². As of 2011 it is the second most populous district of Chhattisgarh (out of 18), after Raipur.[1] The district is home to two important religious sites. The principal Hindu temple, attract pilgrims from all over India. It is also the town of Bhilai Steel Plant.\\
\midrule
   \textbf{ChatGPT}: \textcolor{teal}{Durg district is a district in the Indian state of Chhattisgarh. The district headquarters is located in the city of Durg,} \textcolor{orange}{which is also the largest city in the district. The district has a total area of 2,676 square kilometers} and a population of approximately 3 million people. The economy of Durg district is primarily based on agriculture and \textcolor{teal}{industries}. The district is known for its paddy cultivation.
   %and is one of the largest producers of rice in the state.
\\
% \midrule
% \textbf{ChatGPT-R}: \textcolor{teal}{Durg district is a district located in the Indian state of Chhattisgarh.} the district is known for its contributions to the \textcolor{teal}{industrial development}, cultural competence, social harmony, and meaningful use of resources of the state. \textcolor{teal}{Durg is a symbol of status, prestige}, and glory of Chhattisgarh, with a unique mixture of oldness and modernity, culture-rite, and entrepreneurship. \textcolor{teal}{The district covers an area of 2,238 km²}. \\
\midrule
     \textbf{DKGen}: \textcolor{teal}{Durg district is a district in the state of Chhattisgarh in India, situated on the east bank of the river Shivnath.} \textcolor{red}{[2]} \textcolor{teal}{It is the second most populous district of Chhattisgarh after its capital city, Raipur.}\textcolor{red}{[1]} \textcolor{red}{[3]} \textcolor{teal}{The region is home to some of the biggest and prominent steel plants in the countr.}\textcolor{red}{[4]}  \\
    \bottomrule
    \end{tabular}
     \vspace{-10pt}
    \label{tab:case}
\end{table*}
We carry out a case study to compare the text generated by DKGen and ChatGPT, presented in Table~\ref{tab:case}. Our observations from these cases reveal that: (1) DKGen produces a greater number of sentences supported by the reference or the target text compared to ChatGPT, while a large portion of the text generated by ChatGPT lacks such support (Green text vs. Other); (2) ChatGPT tends to generate more generic text than DKGen, which complicates the assessment of its factual accuracy; (3) DKGen is capable of generating reference marks at the end of each sentence, facilitating the traceability of the source text and thereby enhancing the credibility of the generated content.
\vspace{-5pt}
\section{Conclusion and Limitation}
% In this paper, we present DKGen, a model designed to optimize factual accuracy in text generation through dynamic knowledge selection. Rather than inputting all retrieved reference passages into the generator to create the entire output in a single pass, DKGen divides the text generation process into an iterative process. In this process, DKGen accurately selects reference passages supporting the generation of a single sentence. Once the iterative process is complete, we combine all generated sentences to form the final output. Furthermore, we can enhance the reliability of the output by attaching reference marks at the end of each sentence. We conducted experiments on a public benchmark which demonstrated that DKGen outperforms all baselines.

% Knowledge-enhanced generation is a promising direction for both academic and industrial areas, in which selecting the proper knowledge and determining when to apply it during text generation are urgent issues that require resolution. In this paper, the proposed DKGen provides a potential solution to these issues. 
% Despite its potential benefits, DKGen has a few current limitations: First, it employs heuristic strategies to select the number of reference passages, which may not provide adequate knowledge when synthesizing information from different sources. Second, 
% through human evaluation, the iterative sentence generation yielded less semantic coherence than generating the entire output in one pass. 

In this paper, we introduce DKGen, a model specifically designed to enhance the factual accuracy of text generation by employing dynamic knowledge selection. Instead of incorporating all available reference passages into a single-pass generator, DKGen adopts an iterative process that selects reference passages to support the generation of individual sentences. Upon completion of the iterative process, the generated sentences are combined to produce the final output. To further augment the output's reliability, reference marks can be appended to the end of each sentence. Our experiments on a public benchmark demonstrate that DKGen outperforms all baseline models.

Knowledge-enhanced generation holds big potential applications for both academia and industry. Selecting appropriate knowledge and ascertaining the optimal moment to apply it during text generation remain challenges in need of effective solutions. In this paper, our proposed DKGen model provides a potential solution to address these concerns.
However, DKGen has limitations: Firstly, it relies on heuristic strategies to select the number of reference passages, which may result in insufficient knowledge synthesis from diverse sources. Secondly, human evaluation indicates that the iterative sentence generation approach adopted by DKGen may lead to lower semantic coherence compared to producing the entire output in a single pass. We will address these limitations in the future work.

\bibliography{main}
\bibliographystyle{unsrtnat}

\appendix
\end{document}